\pdfoutput=1
\documentclass[10pt,twocolumn,letterpaper]{article}

\usepackage[pagenumbers]{cvpr} 

\usepackage{graphicx}
\usepackage{amsmath}
\usepackage{amssymb}
\usepackage{booktabs}
\usepackage{adjustbox}
\usepackage{multirow}
\usepackage{color}
\usepackage{bbding}
\usepackage{cite}
\usepackage[pagebackref,breaklinks,colorlinks]{hyperref}

\hypersetup{
	colorlinks=true,
	linkcolor=red,
	filecolor=red,      
	urlcolor=red,
	citecolor=green,
}

\usepackage[capitalize]{cleveref}
\crefname{section}{Sec.}{Secs.}
\Crefname{section}{Section}{Sections}
\Crefname{table}{Table}{Tables}
\crefname{table}{Tab.}{Tabs.}

\def\cvprPaperID{7572} 
\def\confName{CVPR}
\def\confYear{2023}

\begin{document}
\sloppy
\title{NeRF-Gaze: A Head-Eye Redirection Parametric Model for Gaze Estimation}

\author{Pengwei Yin\footnotemark[1], Jiawu Dai\footnotemark[1],  Jingjing Wang, Di Xie, Shiliang Pu \\
Hikvision Research Institute, China \\
\{yinpengwei, daijiawu, wangjingjing9, xiedi,pushiliang\}@hikvision.com
}
 
\maketitle

\renewcommand{\thefootnote}{\fnsymbol{footnote}} 
\footnotetext[1]{These authors contributed equally to this work.} 

\begin{abstract}
Gaze estimation is the fundamental basis for many visual tasks. Yet, the high cost of acquiring gaze datasets with 3D annotations hinders the optimization and application of gaze estimation models. In this work, we propose a novel Head-Eye redirection parametric model based on Neural Radiance Field, which allows dense gaze data generation with view consistency and accurate gaze direction.
Moreover, our head-eye redirection parametric model can decouple the face and eyes for separate neural rendering, so it can achieve the purpose of separately controlling the attributes of the face, identity, illumination, and eye gaze direction.
Thus diverse 3D-aware gaze datasets could be obtained by manipulating the latent code belonging to different face attributions in an unsupervised manner. Extensive experiments on several benchmarks demonstrate the effectiveness of our method in domain generalization and domain adaptation for gaze estimation tasks. 
\end{abstract}

\section{Introduction}
Gaze estimation has been widely applied to human behavior and mental analysis. Gaze estimation with high accuracy can provide strong support for many applications, such as human-computer interaction\cite{8542583,8542505}, augmented reality\cite{10.1145/3084363.3085029} and driver monitoring systems\cite{8326022}.
Furthermore, the appearance-based gaze estimation methods achieved significant performance gains with the recent development of deep learning.

\begin{figure}[ht!]
	\centering
	\includegraphics[width=0.46\textwidth]{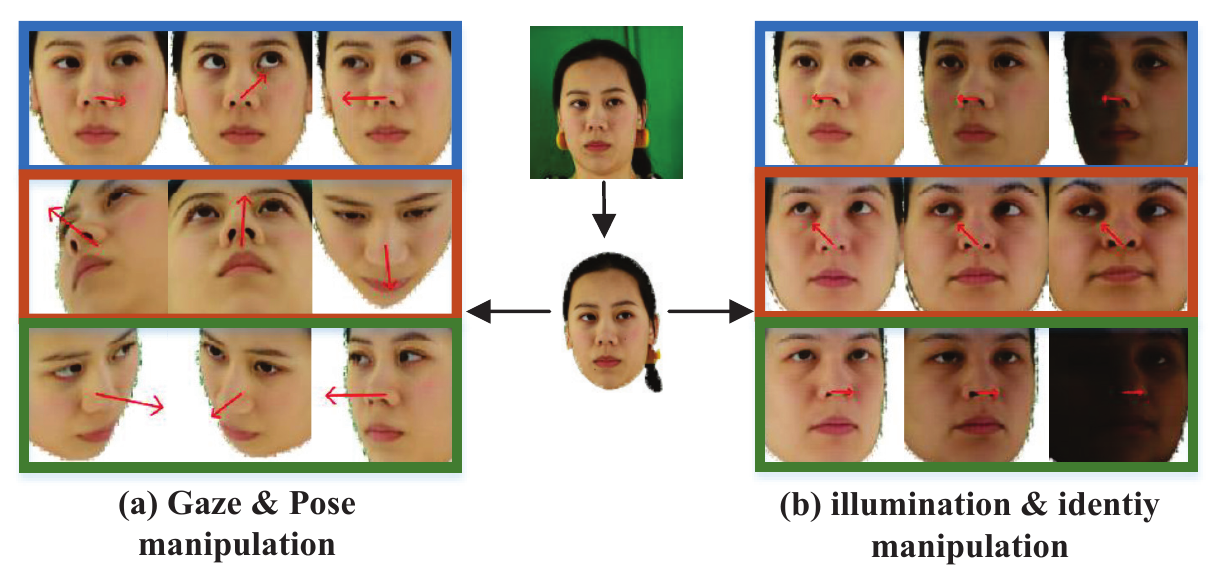}
	\caption{Examples for gaze data generated by our NeRF-Gaze, 
 (a) represents manipulation of gaze direction and pose orientation, 
 $top$: only gaze direction, 
 $middle$: only pose orientation, 
 $bottom$: gaze and pose. 
 While (b) illustrates the facial attribute adjustments, 
 $top$: only illumination, 
 $middle$: only identity, 
 $bottom$: illumination and identity. 
 It is worth noting that our NeRF-Gaze achieves 
 decoupling of gaze direction, head pose, 
 as well as other gaze-irrelevant attributes.}
	\label{tag:nerf_gaze}
\end{figure}

Since gaze data collection and annotation in various conditions for generalizable gaze estimation is an expensive and time-consuming process, getting accurately labeled samples need to build a sophisticated and well-calibrated system. To this end, the research community focuses on the data generation process for benchmarking with a large variation in data attributes.
Previous methods\cite{he2019photo,Yu2019ImprovingFU} use image generation technology to generate eye images in different viewing directions, but they can only generate a patch eye image in a limited head pose. 
Moreover, most methods\cite{sugano2014learning,wood2016learning,he2019photo,Yu2019ImprovingFU} cannot generate the full-face data, yet the other methods\cite{zheng2020self,qin2022learning} can generate the full-face data with low fidelity and inaccurate gaze label.
The difference between the gaze data generation methods aforementioned is detailed in Table \ref{tab:generation_method_contrast}.

\begin{table*}[htp]
	\renewcommand\arraystretch{1.2}
	\caption{Differences in gaze data generation approaches for gaze estimation.}
	\setlength{\tabcolsep}{5mm}
	\centering
	\label{tab:generation_method_contrast}
	\scalebox{1.0}{
		\begin{tabular}{llllll}
			\toprule[2pt]
			Methods& Head Pose& Eye Direction& Face ID& Illumination& Full-Face\\
			\hline
			Sugano \textit{et al.}\cite{sugano2014learning}& \textcolor{green}\CheckmarkBold& \textcolor{red}\XSolidBrush& \textcolor{red}\XSolidBrush& \textcolor{red}\XSolidBrush& \textcolor{red}\XSolidBrush\\
			
			Wood \textit{et al.}\cite{wood2016learning}& \textcolor{green}\CheckmarkBold& \textcolor{green}\CheckmarkBold&  \textcolor{red}\XSolidBrush& \textcolor{green}\CheckmarkBold& \textcolor{red}\XSolidBrush\\
			
			He \textit{et al.}\cite{he2019photo}& \textcolor{red}\XSolidBrush& \textcolor{green}\CheckmarkBold& \textcolor{red}\XSolidBrush& \textcolor{red}\XSolidBrush& \textcolor{red}\XSolidBrush\\
			
			Yu \textit{et al.}\cite{Yu2019ImprovingFU}& \textcolor{red}\XSolidBrush& \textcolor{green}\CheckmarkBold& \textcolor{red}\XSolidBrush& \textcolor{red}\XSolidBrush& \textcolor{red}\XSolidBrush\\
			
			Zheng \textit{et al.}\cite{zheng2020self}& \textcolor{green}\CheckmarkBold& \textcolor{green}\CheckmarkBold& \textcolor{red}\XSolidBrush& \textcolor{red}\XSolidBrush& \textcolor{green}\CheckmarkBold\\
			
			Qin \textit{et al.}\cite{qin2022learning}& \textcolor{green}\CheckmarkBold& \textcolor{red}\XSolidBrush& \textcolor{red}\XSolidBrush& \textcolor{green}\CheckmarkBold& \textcolor{green}\CheckmarkBold\\
			
			Our Method&  \textcolor{green}\CheckmarkBold& \textcolor{green}\CheckmarkBold& \textcolor{green}\CheckmarkBold& \textcolor{green}\CheckmarkBold& \textcolor{green}\CheckmarkBold\\
			
			\bottomrule[2pt]
	\end{tabular}}
\end{table*}

Recently, Neural Radiance Field (NeRF) has become a hot spot for 3D-aware generation tasks due to its inherent differentiable property and high-fidelity rendering. Some pioneering works applied NeRF in head parameterization or face editing. HeadNeRF\cite{hong2021headnerf} utilized NeRF to model the human head, which not only renders head images with novel poses but also supports the manipulation of expressions and appearance of the human face. FENeRF\cite{Sun_2022_CVPR} applied disentangled latent codes to generate facial semantics, which can be used to conduct 3D face editing via GAN inversion. These approaches achieve remarkable face modeling and manipulation results, but none of them render the eye image independently of the face, much less precise control of eye gaze direction.

To solve the issues of previous methods, we propose NeRF-Gaze, a novel gaze data generation method that is based on the neural radiance field. To the best of our knowledge, this is the first attempt to apply NeRF to gaze data generation.
Specifically, in addition to the NeRF model for the face, we also build individual models for the eyes and control the eye gaze by the eye direction vector under the head coordinate system.
Typically, the gaze direction is the coupling of eye direction and head pose, and we merely utilize the eye direction to explicitly control the gaze to achieve a natural disentangling.
Unlike the previous gaze redirection approaches, our method can easily adjust the head pose and gaze direction simultaneously for a given head image. As shown in Figure \ref{tag:nerf_gaze}(a), our NeRF-Gaze is free to render view-consistent head images with specific gaze direction.  


In addition, our NeRF-Gaze also disentangles the gaze direction and other facial attributes, including identity and illumination by decoupling the face and eyes with separate neural rendering. Inspired by \cite{hong2021headnerf}, we regard the shape and appearance transformation factors while fitting a pre-trained nonlinear 3D face morphable model as latent codes for identity and illumination, respectively.
Those latent codes are also utilized as the input of NeRF-Gaze along with the gaze direction.
Thanks to the efficient separate neural rendering, our NeRF-Gaze enables the manipulation of gaze-irrelevant facial factors.
Figure \ref{tag:nerf_gaze}(b) illustrates some examples of identity and illumination transformation.
Given the remarkable performance of gaze redirection and attribute manipulation, our method can be used to generate novel and accurate gaze datasets with specific gaze distributions, identity changes, and illumination conditions, which is practical for downstream domain generalization and domain adaptation gaze estimation tasks. The main contributions of this work can be summarized as follows:
\begin{itemize}
	\item We propose the first NeRF-based method for gaze data generation, which can synthesize novel views of head images and can precisely control the direction of eye gaze with high fidelity.
	\item To control the direction of eye gaze more accurately, we propose a flexible NeRF framework, which can decouple the face and eyes for separate neural rendering, making that parts of the face can be controlled without affecting the other face region.
	\item Extensive experiments demonstrate the proposed NeRF-Gaze can be used to generate high-fidelity gaze data by individually adjusting gaze direction and other different facial factors, and facilitate the downstream gaze estimation task in domain generalization and domain adaptation settings.
\end{itemize}	 
 
\section{Related Works}
\textbf{Gaze Estimation and Adaptation.} Gaze estimation based CNN is currently a hot spot\cite{Zhang2015AppearancebasedGE,Krafka2016EyeTF,Cheng_2018_ECCV,Cheng2020GazeEB}, thanks to the availability of large-scale datasets\cite{zhang2020eth,kellnhofer2019gaze360,RT_GENE,zhang2017mpiigaze} and the rapid development of CNN.
Gaze estimation still has many challenges, such as complex changes in head pose, illumination, etc., as well as subject individual differences.
To alleviate the above issues about domain gaps for gaze estimation, few-shot learning (FSL)\cite{Park2019FewShotAG,Yu2019ImprovingFU} methods learn person-specific gaze models with very few ($\leq$9) labeled samples. Recently, some researchers \cite{Guo_2020_ACCV,liu2021PnP_GA,bao2022generalizing,wang2022contrastive} try to leverage unsupervised domain adaptation (UDA) technique to achieve better performance in the target domain by learning domain invariant features using labeled source domain data and unlabeled target domain data.

\begin{figure*}[ht]
	\centering
	\includegraphics[width=1.0\textwidth]{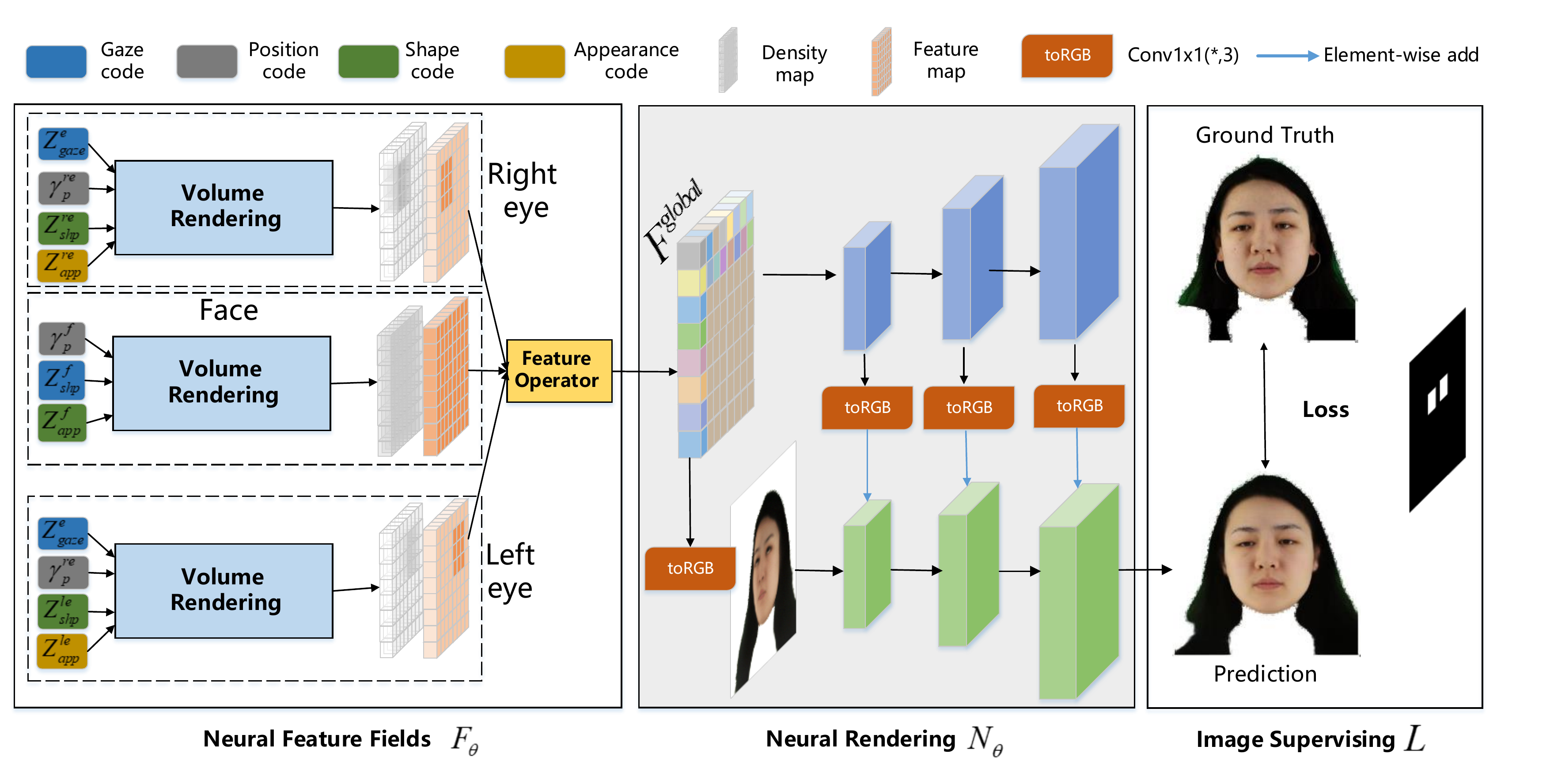}
	\caption{\textbf{Overview of NeRF-Gaze}. In our neural feature fields $F_\theta$, the corresponding feature maps of the face, left and right eyes are obtained by their respective volume rendering modules, and then the feature operator module merges these three feature maps and three density maps to obtain a global face feature map. In the neural rendering $N_\theta$, we render the predicted image $I_{pred}$. Finally, the whole network is trained in a differentiable way by 2D images supervised.}
	\label{tag:framework}
\end{figure*}

\textbf{Gaze Data Generation}
The first gaze data Generation method\cite{sugano2014learning} use a multi-view 3D reconstructed method to generate eye images with different head pose. 
Some works\cite{he2019photo,Yu2019ImprovingFU} use gaze redirection techniques to generate eye images with new gaze direction, but these methods can only be applied in a limited head pose.
More interestingly, some researchers utilize computer
graphics techniques\cite{wood2016learning} to synthesize eye images by building a dynamic eye-region model from head scan geometry.   
More recently, ST-ED\cite{zheng2020self} use an Encoder-Decoder network for redirecting gaze and head pose, which can be also used to generate training data. However, such the GAN method cannot render high fidelity face images for gaze estimation. Extending gaze distribution by monocular 3D face reconstruction \cite{qin2022learning} can not get the accurate gaze label. None of the aforementioned methods can generate high-quality full-face gaze data, and control attributes such as head pose, eye direction, face ID and illumination, etc. These works inspire us to propose a more omniscient and effective gaze data generation method.

\noindent
\textbf{Neural Radiance Field.}
In recent years, NeRF\cite{mildenhall2020nerf} has made great progress in the novel view synthesis area. 
Unlike traditional voxel-based or mesh-based approaches, NeRF represents complex 3D scenes via a continuous implicit function, which can be optimized by merely using sparse viewpoint inputs. 
And subsequent works mainly focus on improving speed\cite{garbin2021fastnerf, deng2022depth, mueller2022instant}, generalizability\cite{trevithick2021grf}, and decreasing the number of views required for the training of NeRF\cite{yu2021pixelnerf}. 
As an efficient 3D-aware generation method, NeRF is also integrated into GANs to achieve flexible semantic control. GIRAFFE\cite{niemeyer2021giraffe} introduces 2D neural rendering modules to accelerate the training of the NeRF-based GANs model.
Control-NeRF\cite{lazova2022control} and Block-NeRF\cite{tancik2022block} can decouple rendering of different objects or scenes, and finally synthesize the global image.
More Recently, HeadNeRF was proposed to parameterize the human head by combining NeRF and 2D neural rendering. Similar to our approach, HeadNeRF also utilizes the latent codes from 3DMM\cite{tran2019towards} to disentangle facial attributes. 
But different from these methods, our NeRF-Gaze individually edits eye feature volume without affecting the overall face image, and enables accurate manipulation of head pose and gaze direction simultaneously, which is beneficial to gaze estimation tasks in application scenarios.

\section{Method}
The framework of our Head-Eye redirection parametric model is shown in Figure \ref{tag:framework}. 
In this work, we propose a novel head-eye redirection parametric model for gaze estimation. Unlike the previous domain adaptation or representation learning methods, we formulate the parametric face representation under the neural radiance field paradigm.

\subsection{Problem Formulation}
HeadNeRF\cite{hong2021headnerf} is a NeRF-based parametric model, which can render a face image I with a given
camera parameter $R^f$ and some semantic codes $Z_{shp}^{f}$ and $Z_{app}^{f}$. It can be formulated as:
\begin{equation}
I = H(R^f, Z_{shp}^{f}, Z_{app}^{f})
\end{equation}
Although HeadNeRF supports editing the head pose, the identity, expression, and appearance of face images,
it cannot independently control the attributes of a local region of the face, such as the gaze direction of the eyes.

Different from HeadNeRF, our proposed NeRF-Gaze can precisely control the gaze direction of the local eye region by decoupling the face and eyes for separate neural rendering.
Our NeRF-Gaze can generate a face image $I$ with specific gaze direction and attributes of face and eyes under the given camera parameters $R$, which includes intrinsics and extrinsics. It could be donated as:
\begin{equation}
	\label{tag:nerf_gaze_gongshi}
	I = H(R^f, Z_{shp}^{f}, Z_{app}^{f}, R^e, Z_{shp}^{e}, Z_{app}^{e},Z_{gaze}^{e})
\end{equation}
where the $Z_{gaze}^{e}$ stands the gaze embedding obtained from the ground truth eye gaze vector via a Fourier feature mapping proposed in \cite{tancik2020fourier}. Following previous works\cite{tran2019towards, hong2021headnerf}, we combine face identity code with expression code as facial geometric shapes code $Z_{shp}^{f}$, and use face illumination code and face texture as $Z_{app}^{f}$. Specifically, $Z_{app}^{e}$ and $Z_{gaze}^{e}$ are the eye attributes, which are derived from face codes.
 
\subsection{Neural Feature Field}
\subsubsection{Latent Codes}
\textbf{Face Latent Codes.}
For the given various images with gaze direction labels, Firstly, we utilize the off-the-shelf segmentation and landmark tools to locate the human heads, which are used to extract latent codes by fitting a pretrained 3DMM model.
Then, we can obtain the global rigid transformation matrix $R_{ex}$, shape code $Z_{shp}^{f}$, appearance code $Z_{app}^{f}$ during the 3DMM fitting progress. We utilize $R_{ex}$ as the extrinsic parameters for a specific image and use normalized virtual intrinsic parameters $R_{in}$ for the whole dataset. To the end, the combined camera parameters $R^f=[R_{ex}, R_{in}]$ are used to define the rays of the face. 

\noindent
\textbf{Eye Latent Codes.}
We decouple the eyes from the head and render them separately. However, the facial latent codes obtained by 3DMM are not suitable for eyes. To this end, we employ a MLP $\phi^s_\theta$ to regress the eye shape code $Z^{re}_{shp}$ and $Z^{le}_{shp}$from face codes. 
\begin{equation}
	Z^{e}_{shp} =\phi^{s}_\theta(Z_{shp}^{f}, Z_{app}^{f})
\end{equation}

Similarly, another MLP $\phi^a_\theta$ regress the right and left eye appearance code $Z^{re}_{app}$ and $Z^{le}_{app}$from face codes.
\begin{equation}
	Z^{e}_{app} =\phi^{a}_\theta(Z_{shp}^{f}, Z_{app}^{f})
\end{equation}
 
Furthermore, we use camera parameters $R$ which
transforms the 3DMM geometry of face to camera coordinate and image coordinate, then we can estimate the position of the right and left eyes in the camera coordinate and image coordinate. Through the above transformation operations, we get the $R^{re}$ and $R^{le}$ of the right and left eyes, and estimate their mask $M^{e}$ including the eye region in the image.

Different from the previous NeRF-based face synthesis methods\cite{hong2021headnerf,Sun_2022_CVPR}, we volume render the face and eyes separately by employing individual 3D volume rendering.

\subsubsection{Volume Rendering}
\begin{figure}[ht]
	\centering
	\includegraphics[width=0.45\textwidth]{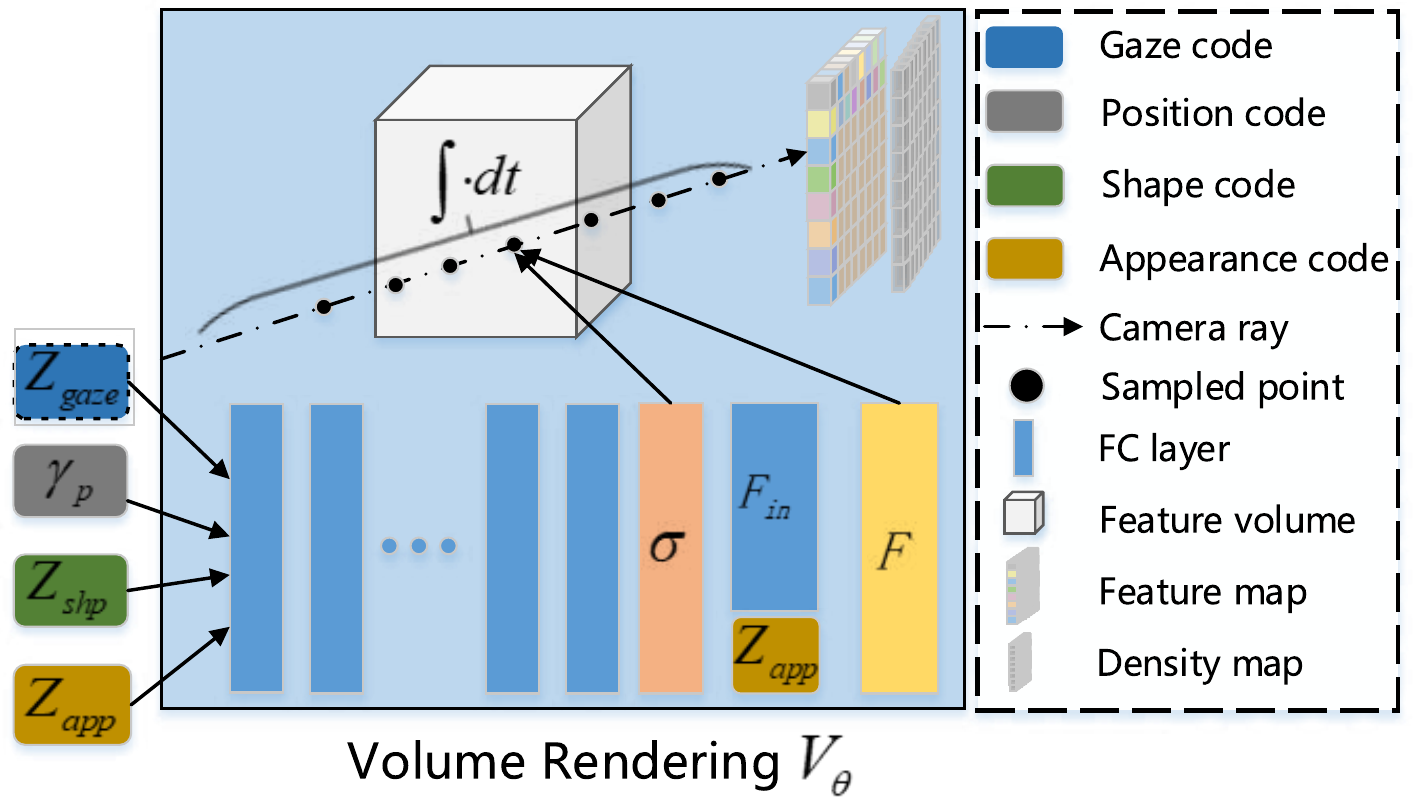}
	\caption{\textbf{Volume Rendering}. We combine the 3D points sampled along camera rays and encode them as position codes, and other codes of facial attributes as input to the MLP-based implicit function. Then we obtain voxels density $\sigma$ and high-dimension features $F$ in the feature volume, which are integrated to predict a low-resolution feature map.}
	\label{tag:nerf_body}
\end{figure}

The volume rendering module is shown in the Figure \ref{tag:nerf_body}, then the NeRF-based scene implicit funtion $h_{\theta}$ with paramemters $\theta$ could be donated as:
\begin{equation}
	(\sigma, F)= h_{\theta}(\gamma_p, Z_{shp}, Z_{app})
\end{equation}
where the $p$ is 3D points $(x, y,z)$ sampled along rays which associate camera parameters $R$, and $\gamma_*$ stands the position encoding, which has been proved a benefit to $h_{\theta}$ to approximate high-frequency function. 

The implicit function $h_{\theta}$ concatenate the embedding of 3D points $\gamma_p$, $Z_{shp}$, and $Z_{app}$ as input, and then output the voxel density $\sigma$ and extra intermediate features $F_{in}$. Next, we combine the appearance attribute $Z_{app}$ with intermediate features to predict a 2D density map $F_{d} \in R^{1\times64\times64}$ and a 2D feature map $F_{f} \in R^{256\times64\times64}$ by classical volume rendering. Therefore, the volume rendering could be formulated as:
\begin{equation}
	(F^f_{d}, F^f_{f})= V_\theta(\gamma_p, Z^f_{shp}, Z^f_{app})
\end{equation}

In addition to the above codes, gaze code is also required to control the eyes for eye rendering. Specially, we utilize eye mask $M^{re}$ and $M^{le}$ to filter out the features of the non-eye region and retain the effective features.
\begin{equation}
(F^e_{d}, F^e_{f})= M^{e}\odot V_\theta(\gamma_p, Z^e_{gaze}, Z^e_{shp}, Z^e_{app})
\end{equation}

\subsubsection{Feature Operator}
We define the feature operator as $O_F$, which uses
the density-weighted mean to combine all features at every pixel$(i,j)$:
\begin{equation}
\begin{split}
\begin{aligned}
F^{global}_{(i,j)} &= \frac{1}{\omega}({F_{f}^f}_{(i,j)} {F_{d}^f}_{(i,j)}
 + {F_{f}^{re}}_{(i,j)} {F_{d}^{re}}_{(i,j)}\\ &+ {F_{f}^{le}}_{(i,j)} {F_{d}^{le}}_{(i,j)})
\end{aligned}
\end{split}
\end{equation}
where $\omega=({F_{d}^f}_{(i,j)} + {F_{d}^{re}}_{(i,j)} + {F_{d}^{le}}_{(i,j)})$. 
In this way, we merge the features of the face and eyes to obtain global features $F^{global}$ which could be formulated as:
\begin{equation}
	F^{global} = O_F(F_{d}^f, {F_{f}^f}; F_{d}^{re}, F_{f}^{re}; F_{d}^{le}, F_{f}^{le})
\end{equation}

\subsection{Neural Rendering}
Inspired by StyleNeRF\cite{gu2021stylenerf}, we decode the feature map $F^{global}$ to obtain the final high-resolution render image $I_{pred} \in R^{512\times512\times3}$ from coarse to fine. The decoder blocks for neural rendering are defined as:
\begin{equation}
	\begin{aligned}
		&Dec(X) = Conv2d(X', K) \\
		&X' = PixShuffle(Repeat(X, 4)+ \phi_{\theta}(X), 2)
	\end{aligned}  
\end{equation}

For any input feature $X \in R^D$, we firstly utilize a 2-layer Conv$1\times1$transformation $\phi_{\theta}:R^D \to R^{4D}$ to get the learnable weights for repeat features, then the aggregated features are upsampled $2\times$ by PixShuffle operation\cite{shi2016real}. The final output of the decoder block is obtained by a 2D Convolutional layer with a fixed blur kernel $K$ like \cite{gu2021stylenerf}. 

\subsection{Image supervising}
We apply several loss functions to optimize the parameters of our Head-Eye redirection model, which include the pixel-level reconstruction loss and the perceptual loss.

\noindent
\textbf{Reconstruction Loss}
For each rendering result from our model, we force the head region and eye region to be consistent with the ground truth image at the pixel level, and the reconstruction loss is donated as:
\begin{equation}
L_{rec} = ||M^{h} \odot(I_{pred}-I_{gt})||^2+||M^{e}\odot(I_{pred}- I_{gt})||^2
\end{equation}
where $M_h$ stands the head mask and $M^{e}$ stands eye mask. 

\noindent
\textbf{Perceptual Loss}
To improve the details of the rendering results, we consider the similarity of the high-level representation associated with human perception. And the perceptual loss on ImageNet pre-trained VGG16 backbone can be written as:
\begin{equation}
\begin{split}
L_{\text {per}}&=\sum_{i}\left\|\psi_{i}(M^{h} \odot I_{pred})-\psi_{i}(M^{h} \odot I_{gt})\right\|^{2}\\
  &+\sum_{i}\left\|\psi_{i}(M^{e} \odot I_{pred})-\psi_{i}(M^{e} \odot I_{gt})\right\|^{2}
\end{split}
\end{equation}
where $\psi_{i}(*)$ is the $i$-th layer's feature map of VGG16 backbone. Therefore, both perception loss and reconstruction loss include face and eye regions.
 
\noindent
\textbf{Total Loss }
In summary, the total loss function applied in our method is:
\begin{equation}
	L_{total} = L_{rec} + L_{per} 
\end{equation}

\section{Experiments}
\subsection{Data Preparation} 
To verify the performance of our method, we conduct extensive experiments on different gaze datasets, which include ETH-XGaze\cite{zhang2020eth}, GazeCapture\cite{krafka2016eye}, Gaze360\cite{kellnhofer2019gaze360}, MPIIFaceGaze\cite{zhang2017mpiigaze} and RT-GENE\cite{RT_GENE}.

\textbf{ETH-XGaze} collects data from 80 subjects with various head poses and gaze direction ranges.
Except for the first 20 subjects training for the NeRF-Gaze model, the middle 45 subjects to train the pre-trained model for gaze estimation, and the last 15 subjects for the in-domain test dataset.

\textbf{GazeCapture} consists of 1474 subjects and over two million frames taken in unconstrained environments and is the largest publicly available gaze dataset.

\textbf{Gaze360} collects 238 participants a wide distribution of participant ages, ethnicity and genders.
We only use the images with face detection annotation, and the training set and test set follow the original settings.

\textbf{MPIIFaceGaze} collects based on the personal computer with a total of 45K images from 15 subjects.

\textbf{RT-GENE} contains 122,531 labeled images with a
total of 15 subjects (9 male, 6 female).

\begin{table*}[htp]
	\renewcommand\arraystretch{1.3}
	\caption{The numerical comparison between different head pose/gaze redirection methods, all metrics are better when lower in value.}
	\setlength{\tabcolsep}{6mm}
	\centering
	\label{tab:generation_contrast}
	\scalebox{1}{
		\begin{tabular}{cccccc}
			\toprule[2pt]
			Methods&  Gaze Redirection& Pose Redirection&  g$\rightarrow$h& h$\rightarrow$g& LPIPS\\ 
			\hline
			StarGAN\cite{choi2018stargan}& 4.602&  3.989&  0.755& 3.067& 0.257\\
			He \textit{et al.}\cite{he2019photo}& 4.617& 1.392&  0.560& 3.925&  0.223\\
			ST-ED\cite{zheng2020self}& 2.195&  0.816&  0.388& 2.072& 0.205\\ 
			Our NeRF-Gaze& \textbf{1.162} & \textbf{0.683} &    \textbf{0.350}& \textbf{1.097}& \textbf{0.116}\\ 
			\bottomrule[2pt]
	\end{tabular}}
	\vspace{-0.2cm}
\end{table*}
 
\subsection{Comparisons to the state-of-the-art baselines}
We compare our method with several SOTA head and gaze redirection approaches, which include StarGAN\cite{choi2018stargan}, \cite{he2019photo}, and ST-ED\cite{zheng2020self}. 
To make a fair comparison, we ensure that the implementation of the experiment is strictly consistent with ST-ED, and the train and test set of GazeCapture are split and processed in exactly the same way.
 \begin{figure}[ht]
 	\vspace{-0.3cm}
	\centering
	\includegraphics[width=0.45\textwidth]{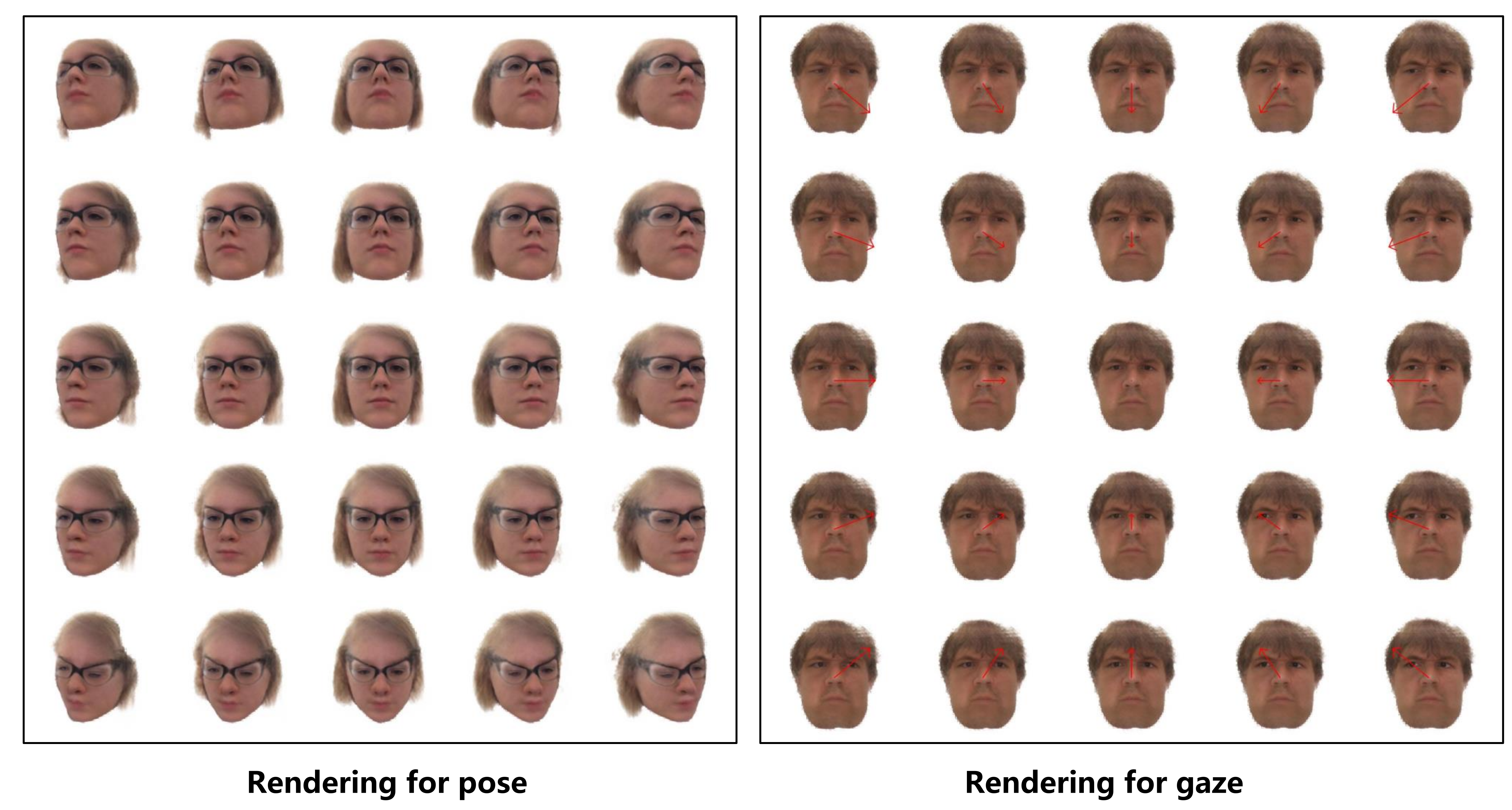}
	\caption{gazecapture-visulization.}
	\label{tag:gazecapture_visulization}
	\vspace{-0.3cm}
\end{figure}

Following ST-ED, we reported the head pose redirection error, gaze redirection error, and the image quality metric LPIPS of different baselines on the GazeCapture (test subset).
Similar to ST-ED, we train a ResNet50-based head estimator and a gaze estimator on the GazeCapture (train subset), and the estimators are used to quantify the angular error between input and redirection images. In addition, we also conduct two variant experiments for disentangle evaluation named $g \rightarrow h$ and $h \rightarrow g$ , and the former measures the head redirection error when merely changing the gaze direction and the latter vice versa.
For the LPIPS metric, we calculated the average score of all image pairs.

To conduct head redirection experiments on our baseline, we replace the camera extrinsic parameters and gaze latent of an input image with the ones of a target image, then the final rendering result as the redirection image as shown in Figure \ref{tag:gazecapture_visulization}.
From the numerical experiments in Table \ref{tab:generation_contrast}, which are all trained on GazeCapture (train subset). We notice that our NeRF-Gaze outperforms all baseline approaches in terms of redirection performance and image generation quality. It proves that our method can generate high-fidelity facial images with accurate head pose and eye direction.
 
\subsection{Effectiveness in Gaze Estimation tasks}
To validate the effectiveness of our redirection model for the gaze estimation tasks, we design extensive experiments across different gaze estimation datasets and backbones.
\begin{figure}[ht]
	\vspace{-0.0cm}
	\centering
	\includegraphics[width=0.45\textwidth]{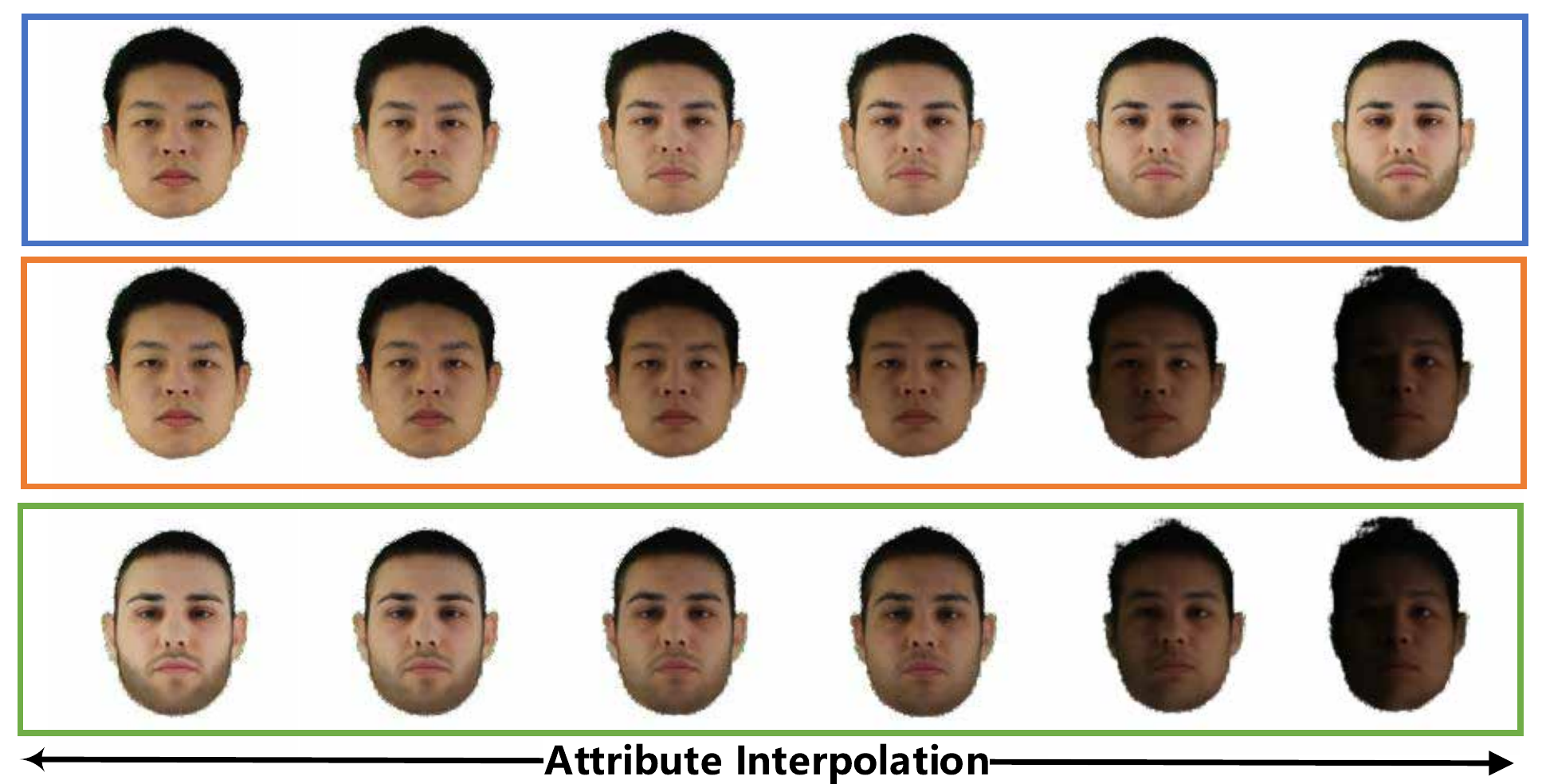}
	\caption{In facial attribute interpolation by our NeRF-Gaze, we conduct linear interpolation on different attributes, top: identity, middle: illumination and bottom: identity and illumination.}
	\vspace{-0.5cm}
	\label{tag:interpolation}
\end{figure}
\subsubsection{Domain Generalization for Gaze Estimation}
The performance of the gaze estimation model can degrade rapidly in cross-dataset scenarios due to gaps in gaze distribution, identity attributes, and illumination between the two domains. Fortunately, our proposed Head-Eye redirection model achieves satisfactory disentanglement of facial properties, which allows us to easily manipulate the face identity as well as the illumination conditions while preserving the head pose and gaze direction. Figure \ref{tag:interpolation} shows the rendering results obtained by interpolating different attributes of our model for a given input image and target image. 

\begin{figure}[ht]
	\centering
	\vspace{-0.1cm}
	\includegraphics[width=0.45\textwidth]{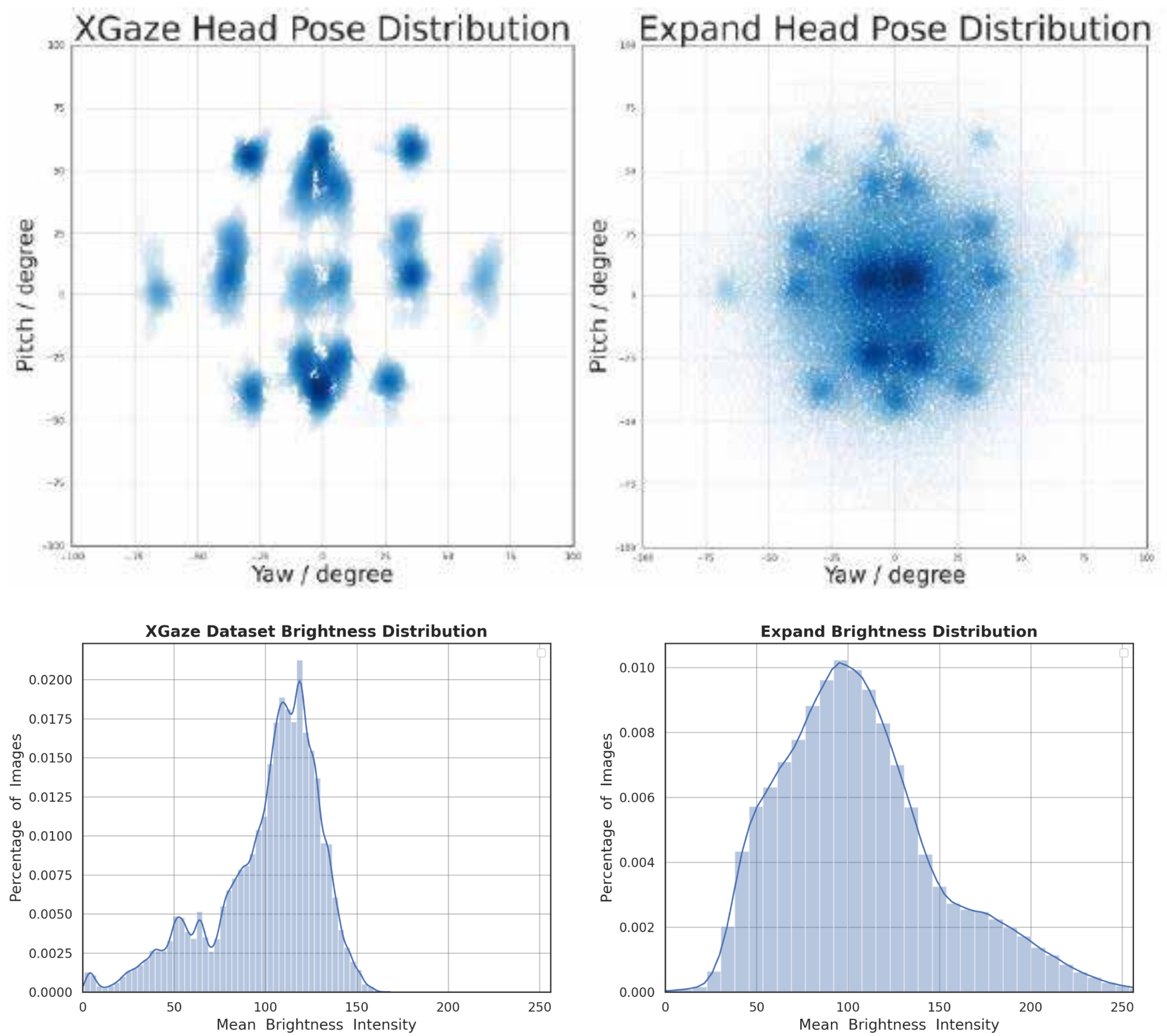}
	\caption{Head pose distribution and brightness distribution for ETH-XGaze and expanding dataset.}
	\label{tag:headpose_ill}
	\vspace{-0.3cm}
\end{figure}

With our NeRF-Gaze, we can extend the number of identities, head pose, and illumination distribution for a specific dataset, which is beneficial for the generalization of the gaze estimation model trained on this dataset. Thus, we conduct dataset extension for ETH-XGaze and Gaze360 by interpolation of head pose and illumination. As shown in Figure \ref{tag:headpose_ill}, we can get an expanded dataset with a denser head pose and smoother illumination distribution for ETH-XGaze. 
 
 \begin{table}[htp]
 	\renewcommand\arraystretch{1.2}
 	\caption{Domain generalization performance for gaze estimation.}
 	\setlength{\tabcolsep}{6mm}
 	\centering
 	\label{tab:narrow_to_wide}
 	\scalebox{1}{
 		\begin{tabular}{cccc}
 			\toprule[2pt]
 			Methods&  E$\rightarrow$M& G$\rightarrow$M \\
 			\hline
 			RT-GENE\cite{RT_GENE}&        $-$&  21.81\\
 			Dilated-Net\cite{chen2018appearance}&   $-$&  18.45\\
 			Full-Face\cite{Zhang2017ItsWA}&  12.35&  11.13\\
 			CA-Net\cite{Cheng2020ACA}&         $-$&  27.13\\
 			ADL\cite{kellnhofer2019gaze360}&         7.23&  11.36\\
 			PureGaze\cite{cheng2022puregaze}&    7.08&   9.28\\
 			\hline
 			Our baseline&          7.9&     9.86\\ 
 			Our NeRF-Gaze&          \textbf{6.98}&     \textbf{7.43}\\ 
 			\bottomrule[2pt]
 	\end{tabular}}
 \end{table}
 
 \begin{table*}[ht!]
 	\renewcommand\arraystretch{1.2}
 	\caption{Comparisons with SOTA Unsupervised Domain Adaptation Methods.}
 	\setlength{\tabcolsep}{6mm}
 	\centering
 	\label{tab:UDA_comparison}  
 	\scalebox{1}{
 		\begin{tabular}{ccccc}
 			\toprule[2pt]
 			Methods&   rely on source-domain?& Target Samples& E$\rightarrow$M& G$\rightarrow$M\\ 
 			\hline
 			ADL\cite{kellnhofer2019gaze360}&  \textcolor{green}\CheckmarkBold&  $>100$&   \textbf{5.48}&  9.70\\
 			DAGEN\cite{Guo_2020_ACCV}& \textcolor{green}\CheckmarkBold& $~500$&   6.16&  6.61\\
 			ADDA\cite{Tzeng2017AdversarialDD}&  \textcolor{green}\CheckmarkBold&   $~500$& 6.33&  8.76\\
 			GVBGD\cite{Cui2020GraduallyVB}& \textcolor{green}\CheckmarkBold&   $~1000$& 6.68&  7.64\\
 			UMA\cite{Cai_2020_CVPR}& \textcolor{green}\CheckmarkBold&   $~100$& 7.52&  8.51\\
 			RSD\cite{chen2021representation}&  \textcolor{green}\CheckmarkBold&   $~100$& 8.74&  9.17\\
 			PNP-GA\cite{liu2021PnP_GA}&  \textcolor{green}\CheckmarkBold&   $100$& 5.53&  6.18\\
 			RUDA\cite{bao2022generalizing}& \textcolor{green}\CheckmarkBold&    $100$& 5.70&  6.20\\
 			CRGA\cite{wang2022contrastive}&  \textcolor{green}\CheckmarkBold&   $100$& 5.68&  \textbf{6.09}\\
 			\hline
 			Our baseline&   \textcolor{red}\XSolidBrush&  0& 7.9&  9.86\\ 
 			Our NeRF-Gaze&   \textcolor{red}\XSolidBrush&  \textbf{$<10$}& 6.36&  7.09\\ 
 			\bottomrule[2pt]
 	\end{tabular}}
 \end{table*}
Finally, we trained two ResNet18-based gaze estimators on the expended datasets, and the test results on MPIIFaceGaze are detailed in Table \ref{tab:narrow_to_wide}.
We totally conduct our cross-domain tasks and denote them as E(ETH-XGaze)$\rightarrow$M(MPIIFaceGaze), G(Gaze360)$\rightarrow$M.
Compared to the performance of previous domain generalization methods, the dataset expanded by NeRF-Gaze brings considerable performance gains in the generalizability of cross-domain gaze estimation tasks.

\subsubsection{Domain Adaptation for Gaze Estimation}
In addition to expanding specific datasets, our NeRF-Gaze is also able to synthesize face images with similar identity and illumination conditions to the target domain from a source domain, and the effect from distribution gaps can be shrunk by fine-tuning the trained gaze model on the synthesized data.
To evaluate the cross-domain generation ability of our method, we conduct cross-domain tasks of E$\rightarrow$M, G$\rightarrow$M.
We trained a ResNet18-based gaze baseline on ETH-XGaze and then fine-tuned the baseline with synthetic gaze data corresponding to each target domain.

Finally, we report the angular gaze error of the baseline model and all variants on the real test split in Table \ref{tab:UDA_comparison}.
We notice that the model fine-tuned on synthetic data outperforms the baseline which is only trained on ETH-XGaze, so this also means that our synthesized target domain data has a positive effect on domain adaptation evaluation. What's more, our results are comparable to most UDA methods but can be independent of source domain data and use very little target domain data $(<10)$ to improve cross-domain performance by using our NeRF-Gaze synthetic data.
This reveals that synthesizing target domain data by our NeRF-Gaze is effective and able to narrow the gaps between the source domain and the target domain.

\subsubsection{Plug Existing Gaze Estimation Methods}
We also apply our gaze data augmented method to different gaze baselines, including FullFace\cite{zhang2017mpiigaze} and Dilated-Net\cite{chen2018appearance}. All the models are trained from scratch and fine-tuned by our synthetic data for ETH-XGaze test split, Gaze360, and RT-GENE. The results are shown in Table \ref{tab:plug}, where we can see that the performance of all baseline models is improved after applying our synthetic gaze data. It indicates that gaze data augmented by our NeRF-Gaze can be utilized as a plug-and-play gaze data augmentation strategy for different gaze baselines.
\begin{table}[htp]
	\renewcommand\arraystretch{1.3}
	\caption{Cross-dataset angular gaze error for plugging our NeRF-Gaze to existing gaze estimation methods.}
	\setlength{\tabcolsep}{4mm}
	\centering
	\label{tab:plug}
	\scalebox{1.0}{
		\begin{tabular}{cccc}
			\toprule[2pt]
			& Gaze360& RT-GENE\\ 
			\hline
			Full-Face\cite{zhang2017mpiigaze}& 26.09& 27.4\\
			Full-Face+Our-data& \textbf{18.87}& \textbf{18.70}\\
			\hline
			Dilated-Net\cite{chen2018appearance}& 27.91& 28.60\\
			Dilated-Net+Our-data& \textbf{19.63}& \textbf{19.04}\\
			\hline
			our baseline& 25.11& 25.89\\ 
			Ours+data& \textbf{18.35}& \textbf{18.97}\\ 
			\bottomrule[2pt]
	\end{tabular}}
\end{table}
\subsection{Ablation Study}

 \begin{figure}[ht]
	\centering
	\includegraphics[width=0.45\textwidth]{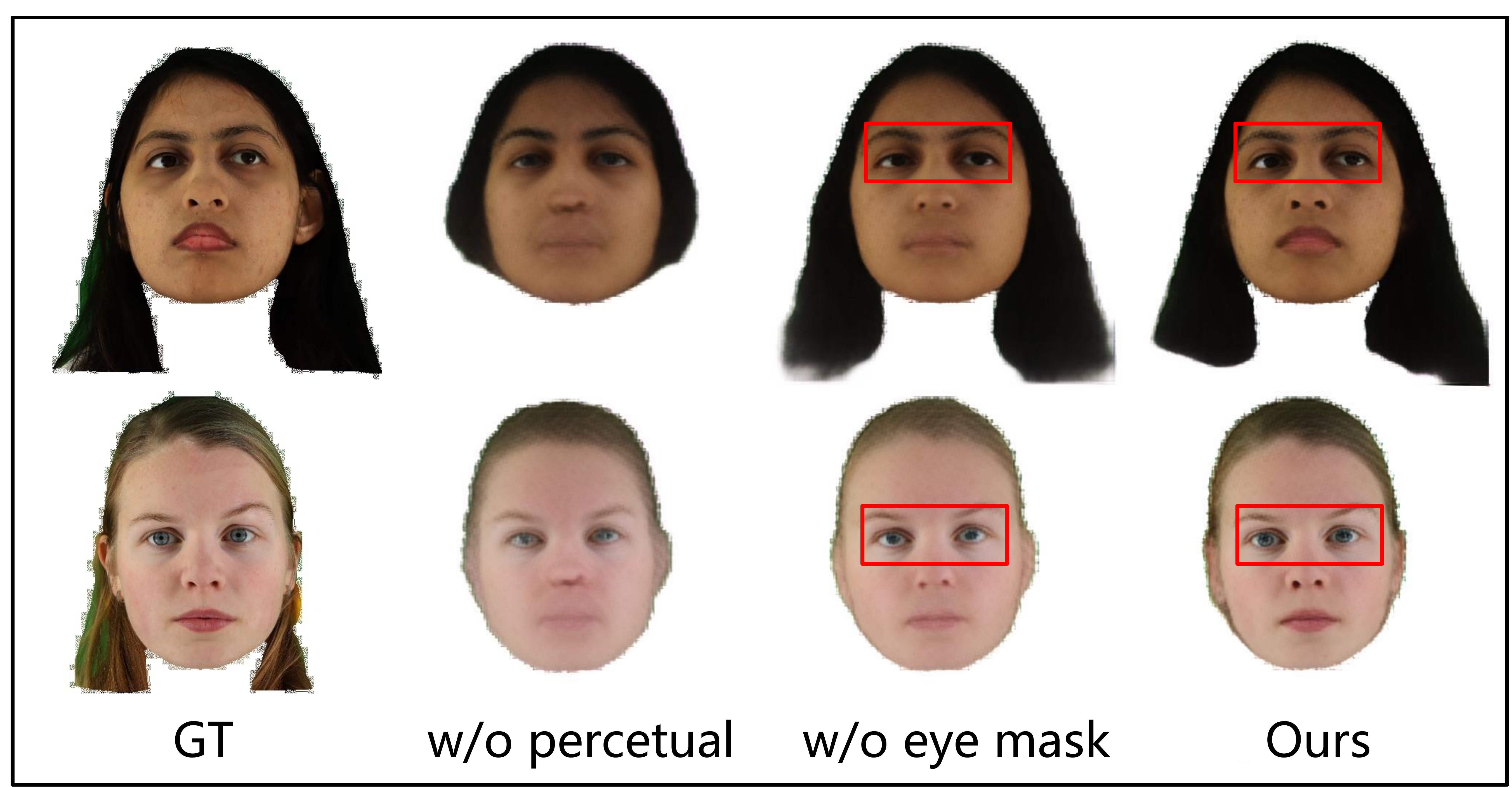}
	\caption{Ablation study on loss functions. Our NeRF-Gaze achieves finer facial details and high fidelity with perceptual loss and eye mask constraints.}
	\label{tag:ablation_for_loss}
	\vspace{-0.3cm}
\end{figure}

\subsubsection{Loss functions}
To verify the effectiveness of our model design, we conduct generation experiments for NeRF-Gaze. Specifically, we remove the different loss constraints for the training of our NeRF-Gaze, the visual comparisons for different settings are illustrated in Figure \ref{tag:ablation_for_loss}. We notice the synthetic images exhibit smooth texture when removing the perceptual loss for both cases in the second column, which is due to the lack of semantic constraints. In the third column of Figure \ref{tag:ablation_for_loss}, the iris boundary inside the eye gets blurred, which indicates the explicit eye mask constraint is the benefit of eye reconstruction. On the contrary, the images generated from our baseline contain distinct textures and fine facial details, which proves the effectiveness of our training strategy. 

\subsubsection{Face attributes}
Furthermore, we conduct cross-domain experiments to investigate the effect of different facial attributes on the generative ability of the proposed NeRF-Gaze. We synthesize gaze data with similar identity and illumination conditions to MPIIFaceGaze, Gaze360 and RT-GENE, and then fine-tune the gaze baseline trained on ETH-XGaze. Table \ref{tab:ablation_crossdomain} elaborates the cross-domain evaluation result, from which we can conclude that both facial attributes are important for the generalization of the gaze dataset.

\begin{table}[htp]
	\renewcommand\arraystretch{1.3}
	\caption{Cross dataset gaze angular error for ablation study on different facial attributes, lower is better.}
	\setlength{\tabcolsep}{1mm}
	\centering
	\label{tab:ablation_crossdomain}
	\scalebox{1}{
		\begin{tabular}{cccc}
			\toprule[2pt]
			&   MPIIFace& Gaze360& RT-GENE\\ 
			\hline
			Baseline& 7.91& 25.11& 25.89\\
			Baseline+ID& 7.18& 21.02& 22.14\\
			Baseline+ID+Illu.& 6.36& 18.35& 18.97\\
			\bottomrule[2pt]
	\end{tabular}}
\vspace{-0.3cm}
\end{table}

\subsubsection{Decoupled rendering}
To demonstrate the effectiveness of our decoupled rendering, we compare the experimental results without decoupling. The method without decoupling is donated as $I=H(R^f, Z_{shp}^{f}, Z_{app}^{f}, Z_{gaze}^e)$, which models the face and eye with a single encoder. Then other training settings are the same as our NeRF-Gaze.

\begin{figure}[ht]
	\centering
	\includegraphics[width=0.45\textwidth]{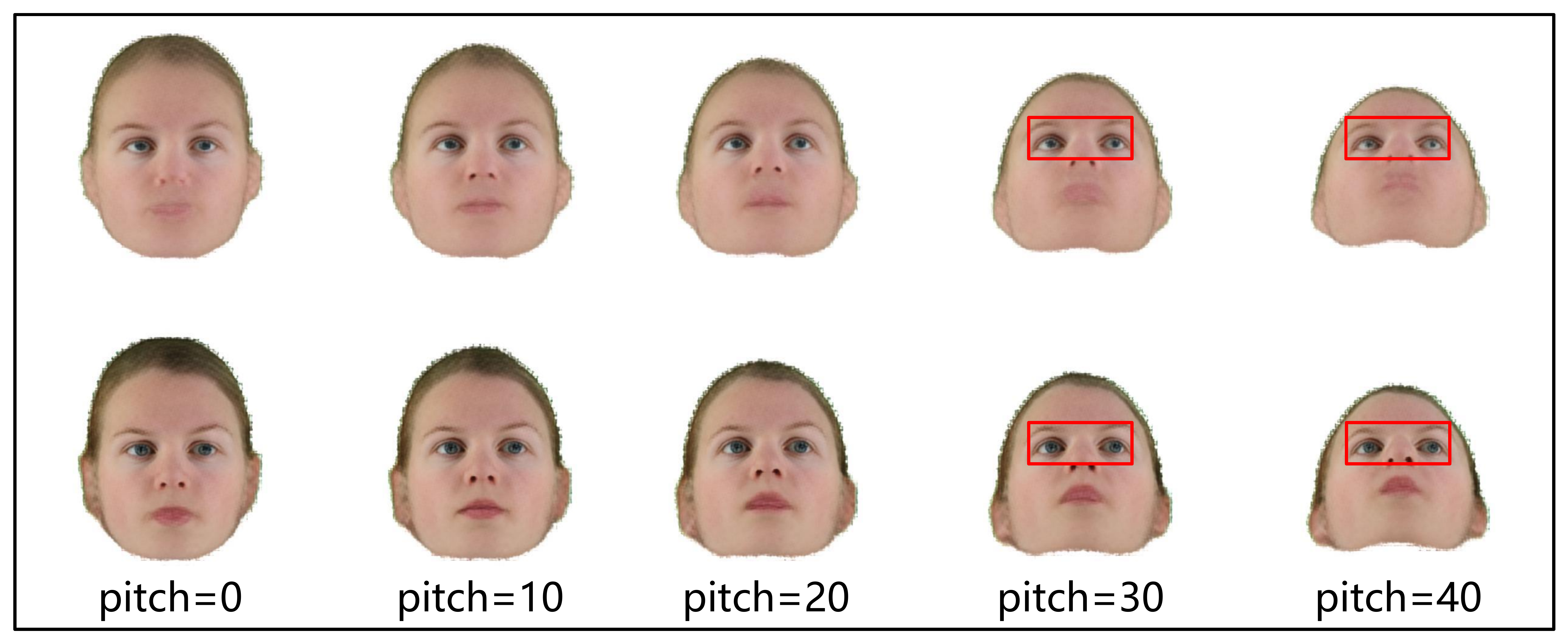}
	\caption{Ablation study on decoupled rendering images.}
	\label{tag:ablation_for_render}
\end{figure}

There is a comparison of visual rendering results, we render a set of images with different head poses as shown on \ref{tag:ablation_for_render}. The first row represents the results without decoupling, and the second row is the result of our NeRF-Gaze. Obviously, when the pitch angle is larger, the eye region of the method without decoupling will become more blurred, while our NeRF-Gaze still renders the eye region perfectly.

In addition, we also conduct redirection and image quality evaluation of the images generated by the two methods in Table \ref{tab:render_contrast}. Our NeRF-Gaze beats the method without decoupling face and eyes on every metric, which also shows that our method could accurately render gaze data.

\begin{table}[htp]
	\renewcommand\arraystretch{1.3}
	\caption{Evaluation for redirection error and image quality.}
	\setlength{\tabcolsep}{1mm}
	\centering
	\label{tab:render_contrast}
	\scalebox{1}{
		\begin{tabular}{cccccc}
			\toprule[2pt]
			Methods&   G.Redir. & P. Redir.&  g$\rightarrow$h& h$\rightarrow$g& LPIPS\\ 
			\hline
			w/o Decoupled& 2.875&  0.830&  0.388& 2.152& 0.125\\ 
			Decoupled& \textbf{1.032} & \textbf{0.653} &    \textbf{0.321}& \textbf{0.977}& \textbf{0.086}\\ 
			\bottomrule[2pt]
	\end{tabular}}
\end{table}

\section{Conclusion}
In this work, we propose NeRF-Gaze, a novel NeRF-based head-eye redirection parametric model for gaze estimation. We utilize a NeRF-like implicit function to parameterize faces and generate low-dimensional image features by volume rendering, which are decoded to high-fidelity faces by the neural rendering modules from coarse to fine. 
Moreover, our NeRF-Gaze builds a flexible NeRF framework, which can decouple the face and eyes for separate neural rendering, making that parts of the face can be controlled without
affecting the other face region, and controls the direction of eye gaze more accurately.
Furthermore, by embedding gaze direction and underlying facial factors into the implicit function, our NeRF-Gaze achieves an effective separation between gaze representations and gaze-irrelevant facial attributes, including identity and illumination.
With our NeRF-Gaze, we can generate diverse and accurate gaze data with specific head pose and gaze direction. The extensive experiments on domain generalization and domain adaptation have demonstrated the effectiveness of our NeRF-Gaze in gaze estimation tasks. 

{\small
	\bibliographystyle{ieee_fullname}
	\bibliography{egbib}

\begin{thebibliography}{10}\itemsep=-1pt

\bibitem{8542583}
Sean Andrist, Xiang~Zhi Tan, Michael Gleicher, and Bilge Mutlu.
\newblock Conversational gaze aversion for humanlike robots.
\newblock In {\em 2014 9th ACM/IEEE International Conference on Human-Robot
  Interaction (HRI)}, pages 25--32, 2014.

\bibitem{bao2022generalizing}
Yiwei Bao, Yunfei Liu, Haofei Wang, and Feng Lu.
\newblock Generalizing gaze estimation with rotation consistency.
\newblock In {\em Proceedings of the IEEE/CVF Conference on Computer Vision and
  Pattern Recognition}, pages 4207--4216, 2022.

\bibitem{Cai_2020_CVPR}
Minjie Cai and Feng Lu.
\newblock Generalizing hand segmentation in egocentric videos with
  uncertainty-guided model adaptation.
\newblock In {\em Proceedings of the IEEE/CVF Conference on Computer Vision and
  Pattern Recognition (CVPR)}, June 2020.

\bibitem{chen2021representation}
Xinyang Chen, Sinan Wang, Jianmin Wang, and Mingsheng Long.
\newblock Representation subspace distance for domain adaptation regression.
\newblock In {\em ICML}, pages 1749--1759, 2021.

\bibitem{chen2018appearance}
Zhaokang Chen and Bertram~E Shi.
\newblock Appearance-based gaze estimation using dilated-convolutions.
\newblock In {\em Asian Conference on Computer Vision}, pages 309--324.
  Springer, 2018.

\bibitem{cheng2022puregaze}
Yihua Cheng and Yiwei Bao.
\newblock Puregaze: Purifying gaze feature for generalizable gaze estimation.
\newblock In {\em Proceedings of the AAAI Conference on Artificial
  Intelligence}, pages 436--443, 2022.

\bibitem{Cheng2020ACA}
Yihua Cheng, Shiyao Huang, Fei Wang, Chen Qian, and Feng Lu.
\newblock A coarse-to-fine adaptive network for appearance-based gaze
  estimation.
\newblock In {\em AAAI}, 2020.

\bibitem{Cheng_2018_ECCV}
Yihua Cheng, Feng Lu, and Xucong Zhang.
\newblock Appearance-based gaze estimation via evaluation-guided asymmetric
  regression.
\newblock In {\em Proceedings of the European Conference on Computer Vision
  (ECCV)}, September 2018.

\bibitem{Cheng2020GazeEB}
Yihua Cheng, Xucong Zhang, Feng Lu, and Yoichi Sato.
\newblock Gaze estimation by exploring two-eye asymmetry.
\newblock {\em IEEE Transactions on Image Processing}, 29:5259--5272, 2020.

\bibitem{choi2018stargan}
Yunjey Choi, Minje Choi, Munyoung Kim, Jung-Woo Ha, Sunghun Kim, and Jaegul
  Choo.
\newblock Stargan: Unified generative adversarial networks for multi-domain
  image-to-image translation.
\newblock In {\em Proceedings of the IEEE conference on computer vision and
  pattern recognition}, pages 8789--8797, 2018.

\bibitem{Cui2020GraduallyVB}
Shuhao Cui, Shuhui Wang, Junbao Zhuo, Chi Su, Qingming Huang, and Qi Tian.
\newblock Gradually vanishing bridge for adversarial domain adaptation.
\newblock {\em 2020 IEEE/CVF Conference on Computer Vision and Pattern
  Recognition (CVPR)}, pages 12452--12461, 2020.

\bibitem{deng2022depth}
Kangle Deng, Andrew Liu, Jun-Yan Zhu, and Deva Ramanan.
\newblock Depth-supervised nerf: Fewer views and faster training for free.
\newblock In {\em Proceedings of the IEEE/CVF Conference on Computer Vision and
  Pattern Recognition}, pages 12882--12891, 2022.

\bibitem{RT_GENE}
Tobias Fischer and Hyung~Jin Chang.
\newblock Rt-gene: Real-time eye gaze estimation in natural environments.
\newblock In {\em Computer Vision -- ECCV 2018}, pages 339--357, Cham, 2018.
  Springer International Publishing.

\bibitem{garbin2021fastnerf}
Stephan~J Garbin, Marek Kowalski, Matthew Johnson, Jamie Shotton, and Julien
  Valentin.
\newblock Fastnerf: High-fidelity neural rendering at 200fps.
\newblock In {\em Proceedings of the IEEE/CVF International Conference on
  Computer Vision}, pages 14346--14355, 2021.

\bibitem{gu2021stylenerf}
Jiatao Gu, Lingjie Liu, Peng Wang, and Christian Theobalt.
\newblock Stylenerf: A style-based 3d aware generator for high-resolution image
  synthesis.
\newblock In {\em International Conference on Learning Representations}, 2022.

\bibitem{Guo_2020_ACCV}
Zidong Guo, Zejian Yuan, Chong Zhang, Wanchao Chi, Yonggen Ling, and Shenghao
  Zhang.
\newblock Domain adaptation gaze estimation by embedding with prediction
  consistency.
\newblock In {\em Proceedings of the Asian Conference on Computer Vision
  (ACCV)}, November 2020.

\bibitem{he2019photo}
Zhe He, Adrian Spurr, Xucong Zhang, and Otmar Hilliges.
\newblock Photo-realistic monocular gaze redirection using generative
  adversarial networks.
\newblock In {\em Proceedings of the IEEE/CVF International Conference on
  Computer Vision}, pages 6932--6941, 2019.

\bibitem{hong2021headnerf}
Yang Hong, Bo Peng, Haiyao Xiao, Ligang Liu, and Juyong Zhang.
\newblock Headnerf: A real-time nerf-based parametric head model.
\newblock In {\em {IEEE/CVF} Conference on Computer Vision and Pattern
  Recognition (CVPR)}, 2022.

\bibitem{kellnhofer2019gaze360}
Petr Kellnhofer, Adria Recasens, Simon Stent, Wojciech Matusik, and Antonio
  Torralba.
\newblock Gaze360: Physically unconstrained gaze estimation in the wild.
\newblock In {\em Proceedings of the IEEE/CVF International Conference on
  Computer Vision}, pages 6912--6921, 2019.

\bibitem{krafka2016eye}
Kyle Krafka, Aditya Khosla, Petr Kellnhofer, Harini Kannan, Suchendra
  Bhandarkar, Wojciech Matusik, and Antonio Torralba.
\newblock Eye tracking for everyone.
\newblock In {\em Proceedings of the IEEE conference on computer vision and
  pattern recognition}, pages 2176--2184, 2016.

\bibitem{Krafka2016EyeTF}
Kyle Krafka, Aditya Khosla, Petr Kellnhofer, Harini Kannan, Suchendra~M.
  Bhandarkar, Wojciech Matusik, and Antonio Torralba.
\newblock Eye tracking for everyone.
\newblock {\em 2016 IEEE Conference on Computer Vision and Pattern Recognition
  (CVPR)}, pages 2176--2184, 2016.

\bibitem{lazova2022control}
Verica Lazova, Vladimir Guzov, Kyle Olszewski, Sergey Tulyakov, and Gerard
  Pons-Moll.
\newblock Control-nerf: Editable feature volumes for scene rendering and
  manipulation.
\newblock {\em arXiv preprint arXiv:2204.10850}, 2022.

\bibitem{liu2021PnP_GA}
Yunfei Liu, Ruicong Liu, Haofei Wang, and Feng Lu.
\newblock Generalizing gaze estimation with outlier-guided collaborative
  adaptation.
\newblock In {\em Proceedings of the IEEE/CVF International Conference on
  Computer Vision}, 2021.

\bibitem{8326022}
Annu~George Mavely, J.~E. Judith, P.~A. Sahal, and Steffy~Ann Kuruvilla.
\newblock Eye gaze tracking based driver monitoring system.
\newblock In {\em 2017 IEEE International Conference on Circuits and Systems
  (ICCS)}, pages 364--367, 2017.

\bibitem{mildenhall2020nerf}
Ben Mildenhall, Pratul~P Srinivasan, Matthew Tancik, Jonathan~T Barron, Ravi
  Ramamoorthi, and Ren Ng.
\newblock Nerf: Representing scenes as neural radiance fields for view
  synthesis.
\newblock In {\em European conference on computer vision}, pages 405--421.
  Springer, 2020.

\bibitem{8542505}
A.~Jung Moon, Daniel~M. Troniak, Brian Gleeson, Matthew~K.X.J. Pan, Minhua
  Zheng, Benjamin~A. Blumer, Karon MacLean, and Elizabeth~A.t Crof.
\newblock Meet me where i’m gazing: How shared attention gaze affects
  human-robot handover timing.
\newblock In {\em 2014 9th ACM/IEEE International Conference on Human-Robot
  Interaction (HRI)}, pages 334--341, 2014.

\bibitem{mueller2022instant}
Thomas M\"uller, Alex Evans, Christoph Schied, and Alexander Keller.
\newblock Instant neural graphics primitives with a multiresolution hash
  encoding.
\newblock {\em ACM Trans. Graph.}, 41(4):102:1--102:15, July 2022.

\bibitem{niemeyer2021giraffe}
Michael Niemeyer and Andreas Geiger.
\newblock Giraffe: Representing scenes as compositional generative neural
  feature fields.
\newblock In {\em Proceedings of the IEEE/CVF Conference on Computer Vision and
  Pattern Recognition}, pages 11453--11464, 2021.

\bibitem{10.1145/3084363.3085029}
Nitish Padmanaban, Robert Konrad, Emily~A. Cooper, and Gordon Wetzstein.
\newblock Optimizing vr for all users through adaptive focus displays.
\newblock In {\em ACM SIGGRAPH 2017 Talks}, SIGGRAPH '17, New York, NY, USA,
  2017. Association for Computing Machinery.

\bibitem{Park2019FewShotAG}
Seonwook Park, Shalini~De Mello, Pavlo Molchanov, Umar Iqbal, Otmar Hilliges,
  and Jan Kautz.
\newblock Few-shot adaptive gaze estimation.
\newblock {\em 2019 IEEE/CVF International Conference on Computer Vision
  (ICCV)}, pages 9367--9376, 2019.

\bibitem{qin2022learning}
Jiawei Qin, Takuru Shimoyama, and Yusuke Sugano.
\newblock Learning-by-novel-view-synthesis for full-face appearance-based 3d
  gaze estimation.
\newblock In {\em Proceedings of the IEEE/CVF Conference on Computer Vision and
  Pattern Recognition}, pages 4981--4991, 2022.

\bibitem{shi2016real}
Wenzhe Shi, Jose Caballero, Ferenc Husz{\'a}r, Johannes Totz, Andrew~P Aitken,
  Rob Bishop, Daniel Rueckert, and Zehan Wang.
\newblock Real-time single image and video super-resolution using an efficient
  sub-pixel convolutional neural network.
\newblock In {\em Proceedings of the IEEE conference on computer vision and
  pattern recognition}, pages 1874--1883, 2016.

\bibitem{sugano2014learning}
Yusuke Sugano, Yasuyuki Matsushita, and Yoichi Sato.
\newblock Learning-by-synthesis for appearance-based 3d gaze estimation.
\newblock In {\em Proceedings of the IEEE conference on computer vision and
  pattern recognition}, pages 1821--1828, 2014.

\bibitem{Sun_2022_CVPR}
Jingxiang Sun, Xuan Wang, Yong Zhang, Xiaoyu Li, Qi Zhang, Yebin Liu, and Jue
  Wang.
\newblock Fenerf: Face editing in neural radiance fields.
\newblock In {\em Proceedings of the IEEE/CVF Conference on Computer Vision and
  Pattern Recognition (CVPR)}, pages 7672--7682, June 2022.

\bibitem{tancik2022block}
Matthew Tancik, Vincent Casser, Xinchen Yan, Sabeek Pradhan, Ben Mildenhall,
  Pratul~P Srinivasan, Jonathan~T Barron, and Henrik Kretzschmar.
\newblock Block-nerf: Scalable large scene neural view synthesis.
\newblock In {\em Proceedings of the IEEE/CVF Conference on Computer Vision and
  Pattern Recognition}, pages 8248--8258, 2022.

\bibitem{tancik2020fourier}
Matthew Tancik, Pratul Srinivasan, Ben Mildenhall, Sara Fridovich-Keil, Nithin
  Raghavan, Utkarsh Singhal, Ravi Ramamoorthi, Jonathan Barron, and Ren Ng.
\newblock Fourier features let networks learn high frequency functions in low
  dimensional domains.
\newblock {\em Advances in Neural Information Processing Systems},
  33:7537--7547, 2020.

\bibitem{tran2019towards}
Luan Tran, Feng Liu, and Xiaoming Liu.
\newblock Towards high-fidelity nonlinear 3d face morphable model.
\newblock In {\em In Proceeding of IEEE Computer Vision and Pattern
  Recognition}, Long Beach, CA, June 2019.

\bibitem{trevithick2021grf}
Alex Trevithick and Bo Yang.
\newblock Grf: Learning a general radiance field for 3d representation and
  rendering.
\newblock In {\em Proceedings of the IEEE/CVF International Conference on
  Computer Vision}, pages 15182--15192, 2021.

\bibitem{Tzeng2017AdversarialDD}
Eric Tzeng, Judy Hoffman, Kate Saenko, and Trevor Darrell.
\newblock Adversarial discriminative domain adaptation.
\newblock {\em 2017 IEEE Conference on Computer Vision and Pattern Recognition
  (CVPR)}, pages 2962--2971, 2017.

\bibitem{wang2022contrastive}
Yaoming Wang, Yangzhou Jiang, Jin Li, Bingbing Ni, Wenrui Dai, Chenglin Li,
  Hongkai Xiong, and Teng Li.
\newblock Contrastive regression for domain adaptation on gaze estimation.
\newblock In {\em Proceedings of the IEEE/CVF Conference on Computer Vision and
  Pattern Recognition}, pages 19376--19385, 2022.

\bibitem{wood2016learning}
Erroll Wood, Tadas Baltru{\v{s}}aitis, Louis-Philippe Morency, Peter Robinson,
  and Andreas Bulling.
\newblock Learning an appearance-based gaze estimator from one million
  synthesised images.
\newblock In {\em Proceedings of the Ninth Biennial ACM Symposium on Eye
  Tracking Research \& Applications}, pages 131--138, 2016.

\bibitem{yu2021pixelnerf}
Alex Yu, Vickie Ye, Matthew Tancik, and Angjoo Kanazawa.
\newblock pixelnerf: Neural radiance fields from one or few images.
\newblock In {\em Proceedings of the IEEE/CVF Conference on Computer Vision and
  Pattern Recognition}, pages 4578--4587, 2021.

\bibitem{Yu2019ImprovingFU}
Yuechen Yu, Gang Liu, and Jean-Marc Odobez.
\newblock Improving few-shot user-specific gaze adaptation via gaze redirection
  synthesis.
\newblock {\em 2019 IEEE/CVF Conference on Computer Vision and Pattern
  Recognition (CVPR)}, pages 11929--11938, 2019.

\bibitem{zhang2020eth}
Xucong Zhang, Seonwook Park, Thabo Beeler, Derek Bradley, Siyu Tang, and Otmar
  Hilliges.
\newblock Eth-xgaze: A large scale dataset for gaze estimation under extreme
  head pose and gaze variation.
\newblock In {\em European Conference on Computer Vision}, pages 365--381.
  Springer, 2020.

\bibitem{Zhang2015AppearancebasedGE}
Xucong Zhang, Yusuke Sugano, Mario Fritz, and Andreas Bulling.
\newblock Appearance-based gaze estimation in the wild.
\newblock {\em 2015 IEEE Conference on Computer Vision and Pattern Recognition
  (CVPR)}, pages 4511--4520, 2015.

\bibitem{Zhang2017ItsWA}
Xucong Zhang, Yusuke Sugano, Mario Fritz, and Andreas Bulling.
\newblock It’s written all over your face: Full-face appearance-based gaze
  estimation.
\newblock {\em 2017 IEEE Conference on Computer Vision and Pattern Recognition
  Workshops (CVPRW)}, pages 2299--2308, 2017.

\bibitem{zhang2017mpiigaze}
Xucong Zhang, Yusuke Sugano, Mario Fritz, and Andreas Bulling.
\newblock Mpiigaze: Real-world dataset and deep appearance-based gaze
  estimation.
\newblock {\em IEEE transactions on pattern analysis and machine intelligence},
  41(1):162--175, 2017.

\bibitem{zheng2020self}
Yufeng Zheng, Seonwook Park, Xucong Zhang, Shalini De~Mello, and Otmar
  Hilliges.
\newblock Self-learning transformations for improving gaze and head
  redirection.
\newblock {\em Advances in Neural Information Processing Systems},
  33:13127--13138, 2020.

\end{thebibliography}
        
}

\end{document}